\documentclass[twoside]{article}

\usepackage{graphicx}
\usepackage{amsthm}
\usepackage{amsmath}
\usepackage{amssymb}

\usepackage{algorithm}
\usepackage{algorithmic}

\usepackage{multirow}
\usepackage{subcaption}
\usepackage{booktabs}
\usepackage{natbib}

\usepackage{makecell}

\usepackage[accepted]{aistats2025}
\newtheorem{definition}{Definition}

\begin{document}

%

%

\twocolumn[

\aistatstitle{CoGS: Model Agnostic Causality Constrained Counterfactual Explanations using goal-directed ASP\thanks{Authors supported by US NSF Grants IIS 1910131, US DoD, and industry grants. 
}}

\aistatsauthor{ Sopam Dasgupta \And Joaqu\'in Arias \And  Elmer Salazar \And Gopal Gupta }

\aistatsaddress{ University of Texas at \\Dallas, Richardson,\\USA 75080 \And  Universidad Rey Juan Carlos,\\ 28933 Móstoles, Madrid, Spain \And University of Texas at \\Dallas, Richardson,\\USA 75080 \And University of Texas at \\Dallas, Richardson,\\USA 75080 } ]

\begin{abstract}


Machine learning models are increasingly used in critical areas such as loan approvals and hiring, yet they often function as black boxes, obscuring their decision-making processes. Transparency is crucial, as individuals need explanations to understand decisions, primarily if the decisions result in an undesired outcome. Our work introduces \textit{CoGS (Counterfactual Generation with s(CASP))}, a model-agnostic framework capable of generating counterfactual explanations for 
classification models.
\textit{CoGS} leverages the goal-directed Answer Set Programming system s(CASP) to compute realistic
and
causally consistent modifications to feature values, accounting for causal dependencies between them. By using \textit{rule-based machine learning algorithms (RBML)}, notably the FOLD-SE algorithm, \textit{CoGS} extracts the underlying logic of a statistical
model to generate counterfactual solutions. By tracing a step-by-step path from an undesired outcome to a desired one, \textit{CoGS} offers interpretable and actionable explanations of the changes required to achieve the desired outcome. We present details of the \textit{CoGS} framework along with its evaluation.

\end{abstract}

\section{Introduction}

Predictive models are widely used in automated decision-making processes like job-candidate filtering and loan approvals. However, these models often function as black boxes, making it difficult to understand their internal reasoning for decision-making. The decisions made by these models can have significant consequences, leading individuals to seek satisfactory explanations, particularly for unfavourable outcomes. Whether automated systems or humans make decisions, this desire for transparency is crucial. Explaining these decisions presents a significant challenge, particularly when users want to know what changes are required to flip an undesired (negative) decision into a desired (positive) one.


In the approach of \cite{wachter}, \textit{counterfactuals} are employed to explain the reasoning behind a prediction by a machine learning model. Additionally, counterfactuals help answer the question: ``What changes should be made to input attributes or features to flip an undesired outcome to a desired one?" as shown by \cite{ref_Byrne_CF}.  Statistical techniques were used to find counterfactuals-based on the proximity of points in the N-dimensional feature space. However, these counterfactual based approaches by \cite{ref_2_ustun}, \cite{alt_karimi}, \cite{ref_mace_1}, and \cite{ref_mace_2} assume feature independence and do not consider the causal dependencies between features. This results in unrealistic counterfactuals. \textit{MINT} by \cite{ref_4_karimi_2} utilized causal relations between features to generate causally consistent counterfactuals. However, these counterfactuals had limited utility as they followed from a set of simultaneous actions,  which may not always be possible in the real world.


We present the \textit{Counterfactual Generation with s(CASP) (CoGS)} framework, which generates counterfactual explanations for any classification model by using  \textit{rule-based machine learning (RBML)} algorithms such as FOLD-SE by \cite{foldse}. Compared to the work by \cite{wachter}, \textit{CoGS} computes causally compliant counterfactuals for any classification model---whether statistical or rule-based---using Answer Set Programming (ASP) and \textit{rule-based machine learning (RBML)} algorithms. Compared to the work by \cite{ref_4_karimi_2}, \textit{CoGS} takes causal dependencies among features into account when computing a step-by-step path to obtaining these counterfactuals. Another novelty of the \textit{CoGS} framework is that it further leverages the FOLD-SE algorithm to automatically discover potential dependencies between features that are subsequently approved by a user. 


\textit{CoGS} models various scenarios (or worlds): the current \textit{initial state}, $i$, represents a negative outcome, while the \textit{goal state}, $g$, represents a positive outcome. A state is represented as a set of feature-value pairs. \textit{CoGS} finds a path from the \textit{initial state} $i$ to the \textit{goal state} $g$ by performing interventions (or transitions), where each intervention corresponds to changing a feature value while accounting for causal dependencies among features. These interventions ensure realistic and achievable changes from state $i$ to $g$. \textit{CoGS} relies on Answer Set Programming (ASP), as described by \cite{gelfond-kahl}, specifically using the goal-directed s(CASP) ASP system by \cite{scasp-iclp2018}. The problem of finding these interventions can be viewed as a planning problem as shown by \cite{gelfond-kahl}. However, unlike the standard planning problem, the interventions that take us from one state to another are not mutually independent.

\section{Background and Related Work}

\subsection{Counterfactual Reasoning}

Explanations help us understand decisions and inform actions. \cite{wachter} advocated using counterfactual explanations (CFE) to explain individual decisions, offering insights on achieving desired outcomes. For instance, a counterfactual explanation for a loan denial might state: If John had \textit{good} credit, his loan application would be approved. This involves imagining alternate (reasonably plausible) scenarios where the desired outcome is achievable.

For a binary classifier given by $f:X \rightarrow \{0,1\}$, we define a set of counterfactual explanations $\hat{x}$ for a factual input $x \in X$ as $\textit{CF}_{f}(x)=\{\hat{x} \in X | f(x) \neq f(\hat{x})\}$. 
This set includes all inputs $\hat{x}$ leading to different predictions than the original input $x$ under $f$.

Various methods were proposed by \cite{wachter}, \cite{ref_2_ustun} and \cite{alt_karimi} that offered counterfactual explanations. However, they assumed feature independence. This resulted in unrealistic counterfactuals as in the real world, causal dependencies exist between features.

\subsection{Causality and Counterfactuals}

Causality relates to the cause-effect relationship among variables, where one event (the cause) directly influences another event (the effect). In the causal framework by \cite{SCM}, causality is defined through \textit{interventions}. An \textit{intervention} involves external manipulation of $P$ (by the \textit{do}-operator), which explicitly changes $P$ and measures the effect on $Q$. 
This mechanism is formalized using \textit{Structural Causal Models (SCMs)}, which represent the direct impact of $P$ on $Q$ while accounting for all confounding factors. \textit{SCMs} allow us to establish causality by demonstrating that an intervention on $P$ leads to a change in $Q$.

In \textit{SCM}-based approaches such as \textit{MINT} by \cite{ref_4_karimi_2}, capturing the downstream effects of interventions is essential to ensure causally consistent counterfactuals. By explicitly modelling counterfactual dependencies, \textit{SCMs} help in generating these counterfactuals. While \textit{MINT} produces causally consistent counterfactuals, it typically proposes a simultaneous set of minimal interventions/actions. When applied together, this set of interventions produces the counterfactual solution. However, executing this set of interventions simultaneously in the real world might not be possible. The order matters as some interventions may depend on others or require different time-frames to implement. Thus, although \textit{MINT} identifies the minimal set of interventions needed to obtain a counterfactual, the order of these interventions is crucial in the real world. Sequence planning is necessary to account for dependencies and practical constraints.

\subsection{Answer Set Programming (ASP)}
\textbf{Answer Set Programming (ASP)} is a paradigm for knowledge representation and reasoning described in \cite{cacm-asp}, \cite{baral}, \cite{gelfond-kahl}. It is widely used in automating commonsense reasoning. ASP inherently supports non-monotonic reasoning that allows conclusions to be retracted when new information becomes available. This is helpful in dynamic environments where inter-feature relationships may evolve, allowing ASP to reason effectively in the presence of incomplete or changing knowledge.
In ASP, we can model the effect of interventions by defining rules that encode the relationship between variables. While ASP does not explicitly use a \textit{do}-operator, it can simulate interventions through causal rules. For example, rules in ASP specify that when $P$ is $TRUE$, $Q$ follows, and similarly, when $P$ is $FALSE$, $Q$ is also $FALSE$: $(P \Rightarrow Q)$ $\wedge$ $(\neg P \Rightarrow \neg Q)$. By changing $P$ (representing an intervention), ASP can simulate the effect of this change in $Q$, thereby capturing causal dependencies similar to \textit{SCMs}. We employ ASP to encode feature knowledge, decision-making, and causal rules, enabling the automatic generation of counterfactual explanations.


To execute ASP programs efficiently, \textbf{s(CASP)} by \cite{scasp-iclp2018} is used. It is a goal-directed ASP system that executes answer set programs in a top-down manner without grounding. Its query-driven nature aids in commonsense and counterfactual reasoning, utilizing proof trees for justification. 
To incorporate negation, s(CASP) adopts \textit{program completion} as described in \cite{baral}, turning ``if'' rules into ``if and only if'' rules: $(P \Rightarrow Q)$ $\wedge$ $(\neg P \Rightarrow \neg Q)$. The ``only if" counterpart corresponding to a rule is called the \textit{dual rule} and is automatically computed by s(CASP). The \textit{dual rules} allow the construction of alternate worlds that lead to the counterfactuals.

Through these mechanisms, ASP in \textit{CoGS} provides a novel framework for generating realistic counterfactual explanations in a \textit{step-by-step} manner.

\subsection{FOLD-SE}

FOLD-SE is an efficient, explainable \textit{rule-based machine learning (RBML)} algorithm for classification developed by \cite{foldse}. 
FOLD-SE generates default rules---a stratified normal logic program---as an \textit{explainable} model from the given input dataset. Both numerical and categorical features are allowed. The generated rules symbolically represent the machine learning model that will predict a label, given a data record. FOLD-SE can also be used for learning rules capturing causal dependencies among features in a dataset. FOLD-SE maintains scalability and explainability, as it learns a relatively small number of rules and literals regardless of dataset size, while retaining good classification accuracy compared to state-of-the-art machine learning methods.

\vspace{-0.1 in}
\section{Overview}
\vspace{-0.07 in}

\subsection{The Problem}
\vspace{-0.05 in}

When an individual (represented as a set of features) receives an undesired negative decision (loan denial), they can seek necessary changes to flip it to a positive outcome. \textit{CoGS} automatically identifies these changes.
For example, if John is denied a loan (\textit{initial state $i$}), \textit{CoGS} models the (positive) scenarios (\textit{goal set $G$}) where he obtains the loan. The query goal `{\tt ?- reject\_loan(john)}' represents the prediction of the classification model regarding whether John’s loan should be rejected, based on the extracted underlying logic of the model used for loan approval. The (negative) decision in the \textit{initial state $i$} should not apply to any scenario in the \textit{goal set $G$}. The query goal `{\tt ?- reject\_loan(john)}' should be {\tt True} in the initial state \(i\) and {\tt False} for all goals in the \textit{goal set $G$}. The problem is to find a series of interventions, namely, changes to feature values, that will take us from $i$ to $g \in G$.
\vspace{-0.15 in}

\subsection{Solution: \textit{CoGS} Approach}\label{sec_CoGS_approach}
\vspace{-0.05 in}

The \textit{CoGS} approach casts the solution as traversing from an \textit{initial state} to a \textit{goal state}, represented as feature-value pairs (e.g., credit score: 600; age: 24). 
In \textit{CoGS}, ASP is used to model interventions and generate a \textit{step-by-step} plan of interventions, tracing how changes propagate from $i$ to $g \in G$. This generation of a plan is a \textit{planning problem}. However, unlike the standard \textit{planning problem}, the interventions that take us from one state to another are not mutually independent. This approach ensures that each interventions respects casual dependencies between variables, offering an explanation of how one action leads to the next. The step-wise approach of \textit{CoGS} contrasts with approaches like \textit{MINT}, which applies all interventions simultaneously and is useful for understanding the dynamic changes in systems with complex causal relationships.

There can be multiple goal states ($G$) that represents the positive outcome. The objective is to turn a negative decision (\textit{initial state i}) into a positive one (\textit{goal state g}) through necessary changes to feature values, so that the query goal `{\tt ?- not reject\_loan(john)}' will succeed for $g \in G$. 

\textit{CoGS} models two scenarios: 1) the negative outcome world (e.g., loan denial, initial state \(i\)), and 2) the positive outcome world (e.g., loan approval, \textit{goal state $g$}) achieved through specific interventions. Both states are defined by specific attribute values (e.g., loan approval requires a credit score $\geq$ 600). \textit{CoGS} symbolically computes the necessary interventions to find a path from $i$ to $g$, representing a flipped decision.

In terms of ASP, the problem of finding interventions can be cast as follows: given a possible world where a query succeeds, compute changes to the feature values (while taking their causal dependencies into account) that will reach another possible world where negation of the query will succeed. \textit{Each of the intermediate possible worlds we traverse must be viable worlds with respect to the rules.} 
We use the s(CASP) query-driven predicate ASP system for this purpose. s(CASP) automatically generates dual rules, allowing us to execute negated queries (such as `{\tt ?- not reject\_loan/1}') constructively.

When the decision query (e.g., `{\tt ?- reject\_loan/1}') succeeds (negative outcome), \textit{CoGS}
finds the state where this query fails (e.g., `{\tt ?- not reject\_loan/1}' succeeds), which constitutes the \textit{goal state $g$}.
In terms of ASP, the task is as follows: given a world where a query succeeds, compute changes to feature values (accounting for causal dependencies) to reach another world where negation of the query will succeed. Each intermediate world traversed must be viable with respect to the rules. We use the s(CASP) query-driven predicate ASP system for this purpose. s(CASP) automatically generates dual rules, allowing us to execute negated queries (such as `{\tt ?- not reject\_loan/1}') constructively.

\textit{CoGS} employs two kinds of actions: 1) Direct Actions: directly changing a feature value, and 2) Causal Actions: changing other features to cause the target feature to change, utilizing the causal dependencies between features. These actions guide the individual from the \textit{$i$} to the \textit{$g$} through intermediate states, suggesting realistic and achievable changes.
Unlike \textit{CoGS}, the approach of \cite{wachter} can output non-viable solutions as it assumes feature independence.

\begin{figure*}[htp]
    \centering    \includegraphics[width=14.5cm]{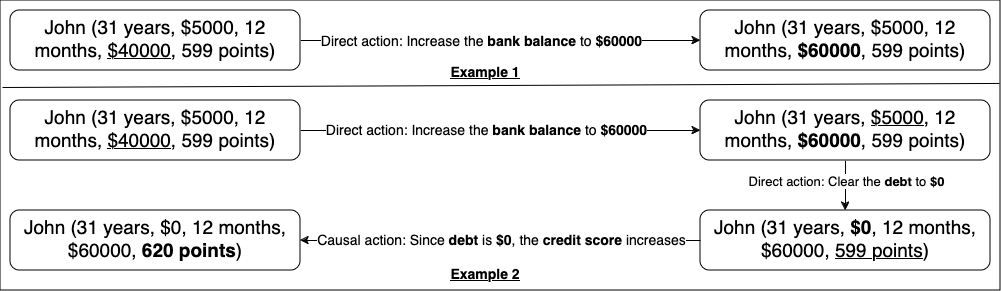}
    \caption{\textbf{Example 1:} John increases his \textit{bank balance} to $\$60000$. \textbf{Example 2:} While the \textit{bank balance} is directly altered by John, altering the \textit{credit score} requires John to directly alter his \textit{debt} obligations first. After clearing his \textit{debt}, the causal effect of having $\$0$ \textit{debt} increases John's \textit{credit score} to $620\ point$.\vspace{-0.2in}}
    \label{fig_example}
\end{figure*}

\smallskip\noindent\textbf{Example 1- Using direct actions to reach the counterfactual state:} \label{eg_1}
Consider a loan application scenario. There are 4 feature-domain pairs: 1) Age: \{1 year,..., 99 years\}, 2) Debt: \{$\$1$, ..., $\$1000000$\}, 3) Bank Balance: \{$\$0$, ..., $\$1\ billion$\} and 4) Credit Score: \{$300\ points, ..., 850\ points$\}. John (31 years, $\$5000$, $\$40000$, $599\ points$) applies for a loan. Based on the extracted underlying logic of the classifier for loan rejection, the bank denies his loan (negative outcome) as his \textit{bank balance} is \textbf{less than $\$60000$}. To get approval (positive outcome), \textit{CoGS} suggests the following:
\textbf{Initial state}: John (31 years, $\$5000$, 12 months, $ \$40000$, $599\ points$) is denied a loan.
\textbf{Goal state}: John (31 years, $\$5000$, 12 months, $\$60000$, $599\ points$) is approved.
\textbf{Intervention}: Increase the \textit{bank balance} to $\$60000$. As shown in Fig. \ref{fig_example}, the direct action flips the decision, making John eligible for the loan.

\smallskip\noindent\textbf{Example 2- Utility of Causal Actions:}
The extracted underlying logic of the classifier for loan rejection produces two rejection rules: 1) individuals with a \textit{bank balance} of \textbf{less than $\$60000$}, and 2) individuals with a \textit{credit score} below $600$. John (31 years, $\$5000$,  $ \$40000$, $599\ points$) is denied a loan (negative outcome) but wants approval (positive outcome). Without causal knowledge, the solution would be: \textbf{Interventions}: 1) Change the \textit{bank balance} to $\$60000$, and 2) the \textit{credit score} to $620\ points$. However, \textit{credit score} cannot be changed directly. To realistically increase the \textit{credit score}, the bank’s guidelines suggest \textbf{having no debt}, indicating a causal dependency between \textit{debt} and \textit{credit score}. \textit{CoGS} provides: \textbf{Initial state}: John (31 years, $\$5000$, 12 months, $ \$40000$, $599\ points$) is denied a loan. \textbf{Goal state}: John (31 years, $\$0$, 12 months, $\$60000$, $620\ points$) is approved for a loan. \textbf{Interventions}: 1) John changes his \textit{bank balance} to $\$60000$, and 2) \textit{reduces his debt} to \$0 to increase his \textit{credit score}. As shown in Figure \ref{fig_example}, clearing the \textit{debt} (direct action) leads to an increase in \textit{credit score}, making John eligible for the loan. Intermediate states (e.g., `$\$5000$ in \textit{debt}' and `$\$0$ in \textit{debt}') represent the path to the goal state. Hence, we demonstrated how leveraging causal dependencies between features leads to realistic desired outcomes through appropriate interventions.


The challenge now is (i) identifying causal dependencies using rule-based machine learning algorithms (FOLD-SE), and (ii) computing the sequence of necessary interventions while avoiding repeating states--- a known issue in the
planning. \textit{CoGS} addresses these by generating the path from the \textit{initial state i} to the counterfactual \textit{goal state g} using the \textit{find\_path} algorithm (Algorithm \ref{alg_path}), explained in Section \ref{sec_alg_path}.

\vspace{-0.13 in}

\section{Methodology}\label{sec_methodology}
\vspace{-0.07 in}

We next outline the methodology used by \textit{CoGS} to generate paths from the initial state (\textit{negative outcome}) to the goal state (\textit{positive outcome}). Unlike traditional planning problems where actions are typically independent, our approach involves \textit{interdependent} actions governed by causal rules $C$. This ensures that the effect of one action can influence subsequent actions, making interventions realistic and causally consistent. 
%
%
Note that the \textit{CoGS} framework uses the FOLD-SE \textit{RBML} algorithm to automatically compute causal dependency rules. These rules have to be either verified by a human, or commonsense knowledge must be used to verify them automatically. This is important, as \textit{RBML} algorithms can identify a correlation as a causal dependency. \textit{CoGS} uses the former approach. We next define specific terms.




\begin{definition}[\textbf{State Space (S)}]\label{defn:S}
$S$ represents all combinations of feature values. For domains \( D_1,..., D_n \) of the features \( F_1,..., F_n\), $S$ is a set of possible states $s$, where each state is defined as a tuple of feature values \( V_1, ..., V_n \). 

\vspace{-0.2in}
\begin{align*}
s\in S\ where\ S = \{ (V_1,V_2,...,V_n)\ | \\
V_i \in D_i,\ for\ each\ i\ in\ 1,...,n \}     
\end{align*}
\end{definition}

\noindent \textit{E.g., an individual John: \textit{s = $(31$ years, $\$5000$, $\$40000$, $599$ points$)$}, where\ $s\in S$.}

\begin{definition}[\textbf{Causally Consistent State Space ($S_C$)}]\label{defn:S_C}
$S_C$ is a subset of $S$ where all causal rules are satisfied. $C$ represents a set of causal rules over the features within a state space $S$. Then, $\theta_C: P(S) \rightarrow P(S)$ (where $P(S)$ is the power set of S) is a function that defines the subset of a given state sub-space $S'\subseteq S$ that satisfy all causal rules in C.

\vspace{-0.1 in}
\begin{equation*}
\theta_C(S') = \{ s \in S' \mid s\ satisfies\ all\ causal\ rules\ in\ C\}
\end{equation*}
\begin{equation*}
S_C = \theta_C(S) 
\end{equation*}


\noindent 
E.g., causal rules state that if $debt$ is $0$, the credit score should be above 599, then instance $s_1$ = (31\ years,\ \$0,\ \$40000,\ 620\ points) is causally consistent, and instance $s_2$ = (31\ years,\ \$0,\ \$40000,\ 400\ points) is causally inconsistent. 

\end{definition}

\noindent 
In a traditional planning problem, allowed actions in a given state are independent, i.e., the result of one action does not influence another. In \textit{CoGS}, causal actions are interdependent, governed by $C$. 

\begin{definition}[\textbf{Decision Consistent State Space ($S_Q$)}]\label{defn:S_Q}
%
$S_Q$ is a subset of $S_C$ where all decision rules are satisfied. $Q$ represents a set of rules that compute some external decision for a given state. $\theta_Q: P(S) \rightarrow P(S)$ is a function that defines the subset of the causally consistent state space $S'\subseteq S_C$ that is also consistent with decision rules in $Q$:
\begin{equation*}
\theta_Q(S') = \{ s \in S' \mid s\ satisfies\ any\ decision\ rule\ in\ Q\} \label{defn:3_1}
\end{equation*}

Given $S_C$ and $\theta_Q$, we define the decision consistent state space $S_Q$ as 
\vspace{-0.055 in}
\begin{equation*}
S_Q = \theta_Q(S_C) = \theta_Q(\theta_C(S)) 
\end{equation*}
\vspace{-0.03 in}
\noindent E.g., an individual John whose loan has been rejected: \textit{s = ($31$ years, $\$0$, $\$40000$, $620$ points)}, where\ $s\in S_Q$.

\end{definition}

\begin{definition}[\textbf{Initial State ($i$)}]\label{defn:I}
$i$ is the starting point with an undesired outcome. Initial state $i$ is an element of the \textit{causally consistent state space} $S_C$
\begin{equation*}
i \in S_C 
\end{equation*}

\noindent For example, $i = (31\ years,\ \$0,\ \$40000,\ 620\ points)$
\end{definition}

\begin{definition}[\textbf{Actions}]\label{defn:action}
The set of actions $A$ includes all possible interventions (actions) that can transition a state from one to another within the state space. Each action $a\in A$ is defined as a function that maps $s$ to a new state $s'$. 
\begin{equation*}
    a: S\rightarrow S\ \mid where\ a\in A
\end{equation*}
\vspace{-0.25in}

Actions are divided into: 1) Direct Actions: Directly change the value of a single feature of a state $s$, e.g.,  Increase bank balance from $\$40000$ to $\$60000$. 2) Causal Actions: Change the value of a target feature by altering related features, based on causal dependencies. It results in a causally consistent state with respect to C, e.g., reduce debt to increase the credit score. 
\end{definition}

\begin{definition}[\textbf{Transition Function}]\label{defn:delta}
A transition function $\delta: S_C \times A \rightarrow S_C$ maps a causally consistent state to the set of allowable causally consistent states that can be reached in a single step, and is defined as: 
\vspace{-0.05in}
  \[\delta(s,a) = \left\{\begin{array}{l}
     a(s) \textit{ if } a(s) \in S_C\\
     \delta(a(s),a') \textit{ with } a\in A, a'\in A \textit{, otherwise}
  \end{array}\right.\]

\noindent $\delta$ models a function that repeatedly takes actions until a causally consistent state is reached. In \textbf{example 1}, $\delta$ suggests  changing the \textit{bank balance} from $\$40000$ to $ \$60000$: 
$\delta(31\ years,\$5000,\$40000,599) = $ $(31\ years,\$5000,\$60000,599)$


\end{definition}

\begin{definition}[\textbf{Counterfactual Generation (CFG) Problem}]\label{problem_statement}
A counterfactual generation (CFG) problem is a 4-tuple $(S_C,S_Q,I,\delta)$ where $S_C$ is causally consistent state space, $S_Q$ is the decision consistent state space, $I\in S_C$ is the initial state, and $\delta$ is a transition function.
\end{definition}

\begin{definition}[\textbf{Goal Set}]\label{defn:G}
The goal set $G$ is the set of desired outcomes that do not satisfy the decision rules $Q$. For the Counterfactual Generation (CFG) problem $(S_C,S_Q,I,\delta)$, $G \subseteq S_C$:
\begin{equation*}
G = \{ s \in S_C| s\not\in S_Q\} 
\end{equation*}
$G$ includes all states in $S_C$ that do not satisfy $S_Q$. For \textbf{example 1}, an example goal state $g\in G$ is $g = (31\ years,\ \$5000,\ \$60000,\ 599\ points).$

\end{definition}

\begin{definition}[\textbf{Solution Path}]\label{solution_path}
A solution to the problem $(S_C,S_Q,I,\delta)$ with Goal set $G$ is a path:
\begin{align*}
    &s_0, s_1, \dots, s_m \ \text{where} \ s_j \in S_C \ \text{for all} \ j \in \{0, \dots, m\}, \\
    &such\ that\ s_0 = I; s_m = G; s_0,...,s_{m-1} \not\in G;\\
    &s_{i+1} \in \delta(s_i) \ \text{for} \ i \in \{0, \dots, m-1\}
\end{align*}

For \textbf{example 1}, individuals with less than $\$60000$ in their account are ineligible for a loan, thus the state of an ineligible individual $s\in S_Q$ might be $s = (31\ years,\$5000,\$40000,599\ points)$. The goal set has only one goal state $g\in G$ given by $s = (31\ years,\$5000,\$60000,599\ points)$. The path from $s$ to $g$ is \{\textit{(31\ years,\$5000,\$40000,599\ points)}$\rightarrow Direct\rightarrow $\textit{(31\ years,\$5000,\$60000,599\ points)}$\}$. Here, the path has only 2 states as only changing the \textit{bank balance} to be $\$60000$ is needed to reach the goal state.
\vspace{-0.1 in}
\end{definition}


\begin{algorithm}[h!]
\caption{\textbf{extract\_logic}: Extract the underlying logic of the classification model}
\label{alg_underlying_logic}
\begin{algorithmic}[1]
    \REQUIRE Original Classification model $M$, Data $H$, \textit{RBML} Algorithm \textit{R}:

    \IF{ $M$ is rule-based}
    \STATE // Decision Rules are the rules of \textit{M}
    \STATE Set \textit{Q} = \textit{M}
    \ELSE
        \IF{ $M$ is statistical}
        \STATE // For input data \textit{H}, predict the labels as \textit{V}
        \STATE Set \textit{V} = \textbf{predict(}\textit{M(H)}\textbf{)}
        \STATE // Use the \textit{H} and \textit{V} to train \textit{R}
        \STATE Set \textit{R} = \textbf{train(}\textit{R(H,V)}\textbf{)}
        \STATE // Decision Rules are the rules of \textit{R}
        \STATE Set \textit{Q} = \textit{R}
        \ENDIF
    \ENDIF
    \STATE Return \textit{Q}

\end{algorithmic}
\end{algorithm}

\vspace{-0.1 in}

\subsection{Algorithm to Extract Decision Rules}

We first describe the algorithm for extracting the underlying logic in the form of rules for the classification model that provides the undesired outcome. By using this extracted logic or rules, we can generate a path to the counterfactual solution $g$.

The function `\textit{\textbf{extract\_logic}}' extracts the underlying logic of the classification model used for decision-making. Our \textit{CoGS} framework applies specifically to tabular data, so any classifier handling tabular data can be used. Algorithm \ref{alg_underlying_logic} provides the pseudocode for `\textit{\textbf{extract\_logic}}', which takes the original classification model $M$, input data $H$, and a \textit{RBML} algorithm $R$ as inputs and returns $Q$, the underlying logic of the classification model.

$Q$ represents the decision rules responsible for generating the undesired outcome. The algorithm first checks if the classification model $M$ is rule-based. If yes, we set $Q=M$ and return Q. Otherwise, the corresponding labels for the input data are predicted using $M$. These predicted labels, along with the input data $H$, are then used to train the \textit{RBML} algorithm $R$. The trained \textit{RBML} algorithm represents the underlying logic of $M$, and we set $Q=R$, returning $Q$ as the extracted decision rules.

\vspace{-0.09 in}
\subsection{Algorithm to Obtain the Counterfactual}
\vspace{-0.05 in}

We next describe our algorithm to find the goal states and compute the solution paths. The algorithm makes use of the following functions: (i) \textbf{not\_member}: checks if an element is: \textit{a}) \textbf{not} a member of a list, and \textit{b}) Given a list of tuples, \textbf{not} a member of any tuple in the list. (ii) \textbf{drop\_inconsistent}: given a list of states [$s_0,...,s_k$] and a set of Causal rules $C$, it drops all the inconsistent states resulting in a list of consistent states with respect to $C$. (iii) \textbf{get\_last}: returns the last member of a list. (iv) \textbf{pop}: returns the last member of a list. (v) \textbf{is\_counterfactual}: returns {\tt True} if the input state is a causally consistent counterfactual solution. (vi) \textbf{intervene}: performs interventions/ makes changes to the current state through a series of actions and returns a list of visited states. The interventions are causally compliant. Further details are available in the supplement.

\begin{algorithm}[h!]
\caption{\textbf{find\_path}: Obtain a path to the counterfactual state}
\label{alg_path}
\begin{algorithmic}[1]
    \REQUIRE Initial State $(I)$, States $S$, Causal Rules $C$, Decision Rules $Q$, \textit{is\_counterfactual}, \textit{intervene}, Actions $a\in A$:

    \STATE\label{PS21} Create an empty list \textit{visited\_states} that tracks the list of states traversed (so that we avoid revisiting them).
    \STATE Append ($i$, [~]) to \textit{visited\_states} 
    \WHILE{\textit{is\_counterfactual(get\_last(visited\_states),C,Q)}\ is\ $FALSE$ }
    \STATE Set \textit{visited\_states=intervene(visited\_states,C,A)}
    \ENDWHILE
    \STATE \textit{candidate\_path = drop\_inconsistent(visited\_states)}
    \STATE Return \textit{candidate\_path}

\end{algorithmic}
\end{algorithm}

\vspace{-0.1 in}
\subsubsection{Find Path:}\label{sec_alg_path}
%
Function `\textit{\textbf{find\_path}}' implements the Solution Path $P$ of Definition \ref{solution_path}. Its purpose is to find a path to the counterfactual state. Algorithm \ref{alg_path} provides the pseudo-code for `\textit{\textbf{find\_path}}', which takes as input an Initial State $i$, a set of Causal Rules $C$, decision rules $Q$, and actions $A$. It returns a path to the counterfactual state/goal state $g\in G$ for the given $i$ as a list `\textit{visited\_states}'. Unrealistic states are removed from `\textit{visited\_states}' to obtain a `\textit{candidate\_path}'.

Initially, $s=i$. The function checks if the current state $s$ is a counterfactual. If $s$ is already a counterfactual, `\textit{\textbf{find\_path}}' returns a list containing $s$. If not, the algorithm moves from $s=i$ to a new causally consistent state $s'$ using the `\textit{\textbf{intervene}}' function, updating `\textit{visited\_states}' with $s'$. It then checks if $s'$ is a counterfactual using `\textbf{\textit{is\_counterfactual}}'. If {\tt True}, the algorithm drops all inconsistent states from `\textit{visited\_states}' and returns the `\textit{candidate\_path}' as the path from $i$ to $s'$. If not, it updates `$current\_state$' to $s'$ and repeats until reaching a counterfactual/goal state $g$. The algorithm ends when `\textbf{\textit{is\_counterfactual}}' is satisfied, i.e., $s'=g\mid\ where\ g \in G$.

\vspace{-0.17 in}
\begin{equation}
\begin{split}
   s'\in G \mid\ by\ definition \\
   s_0,...,s_k,s' \mid s_0,...,s_k: \not\in G
\end{split}
\end{equation}

\subsubsection{Discussion:} A few points should be highlighted: \textbf{(i)} 
%
Certain feature values may be immutable or restricted, such as \textit{age} cannot decrease or \textit{credit score} cannot be directly altered. To respect these restrictions, we introduce \textit{plausibility constraints}. These constraints apply to the actions in our algorithms, ensuring realistic changes to the features. Since they do not add new states but restrict reachable states, they are represented through the set of available actions in Algorithms \ref{alg_path},  
\textbf{(ii)}
Similarly, \textit{CoGS} has the ability to specify the path length for candidate solutions. Starting with a minimal path length of 1, \textit{CoGS} can identify solutions requiring only a single change. If no solution exists, \textit{CoGS} can incrementally increase the path length until a solution is found. This ensures that the generated counterfactuals are both \textbf{minimal and causally consistent}, enhancing their practicality and interpretability. This is achieved via constraints on path length.

\subsection{Soundness}
\vspace{-0.05 in}

    

    
\begin{definition}[CFG Implementation]\label{defn:impl_cfg}
When Algorithm \ref{alg_path} is executed with the inputs: Initial State $i$ (Definition \ref{defn:I}), States Space $S$ (Definition \ref{defn:S}), Set of Causal Rules $C$ (Definition \ref{defn:S_C}), Set of Decision Rules $Q$ (Definition \ref{defn:S_Q}), and Set of Actions $A$ (Definition \ref{defn:action}), a CFG problem $(S_C,S_Q,I,\delta)$ (Definition \ref{problem_statement}) with causally consistent state space $S_C$ (Definition \ref{defn:S_C}), Decision consistent state space $S_Q$ (Definition \ref{defn:S_Q}), Initial State $i$ (Definition \ref{defn:I}), the transition function $\delta$ (Definition \ref{defn:delta}) is constructed. 
    
\end{definition}


\begin{definition}[\textit{Candidate path}]\label{defn:candidate_path}
Given the counterfactual $(S_C,S_Q,I,\delta)$ constructed from a run of  algorithm \ref{alg_path}, the return value (candidate path) is the resultant list obtained from removing all elements containing states $s'\not\in S_C$.     
\end{definition}

\noindent Definition \ref{defn:impl_cfg} maps the input of Algorithm \ref{alg_path} to a \textit{CFG problem} (Definition \ref{problem_statement}). \textit{Candidate path} maps the result of Algorithm \ref{alg_path} to a possible solution (Definition \ref{solution_path}) of the corresponding CGF problem. From Theorem 1 \textit{(proof in supplement)}, the \textit{candidate path} (Definition \ref{defn:candidate_path}) is a solution to the corresponding \textit{CFG problem} implementation (Definition \ref{defn:impl_cfg}).

\noindent \textit{Theorem 1} 
Soundness: Given a CFG $\mathbb{X}$=$(S_C,S_Q,I,\delta)$, constructed from a run of Algorithm \ref{alg_path} \& a corresponding candidate path $P$, $P$ is a solution path for $\mathbb{X}$. Proof is provided in the supplement.

\vspace{-0.1 in}
\begin{table}[h!]
\centering
\fontsize{8}{8}\selectfont



\begin{tabular}{|@{}c@{}|@{}c@{}|@{}c@{}|@{}c@{}|@{}c@{}|@{}c@{}|@{}c@{}|}

\hline
\textbf{\makecell{Dataset}} & \textbf{Model} & \textbf{\makecell{Fid.\\(\%)}} & \textbf{ \makecell{Acc.\\(\%) }} & \textbf{ \makecell{Prec.\\(\%) }} & \textbf{ \makecell{Rec.\\(\%) }} & \textbf{ \makecell{F1.\\(\%) }} \\ 
\hline
\multirow{8}{*}{Adult} & DNN & N/A & \makecell{85.57\\$\pm$0.38} & \makecell{85.0\\$\pm$0.63} & \makecell{85.8\\$\pm$0.40} & \makecell{85.0\\$\pm$0.63} \\ 
\cline{3-7}
&FOLD-SE[DNN] & \makecell{93.16\\$\pm$0.6} & \makecell{84.2\\$\pm$0.28} & \makecell{83.4\\$\pm$0.49} & \makecell{84.2\\$\pm$0.4} & \makecell{83.4\\$\pm$0.49}\\ 
\cline{2-7}
&GBC & N/A &\makecell{86.45\\$\pm$0.33}& \makecell{85.8\\$\pm$0.4} & \makecell{86.6\\$\pm$0.49} & \makecell{85.8\\$\pm$0.4} \\ 
\cline{3-7}
& FOLD-SE[GBC] &  \makecell{95.94\\$\pm$0.67} & \makecell{85.24\\$\pm$0.23} & \makecell{84.6\\$\pm$0.49} & \makecell{85.2\\$\pm$0.4} & \makecell{84.2\\$\pm$0.4} \\ 
\cline{2-7}
& RF & N/A & \makecell{85.60\\$\pm$0.27} & \makecell{85.0\\$\pm$0.0} & \makecell{85.6\\$\pm$0.49} & \makecell{85.2\\$\pm$0.4} \\
\cline{3-7}
& FOLD-SE[RF] & \makecell{90.27\\$\pm$0.3}  & \makecell{84.37\\$\pm$0.23} & \makecell{83.4\\$\pm$0.49} & \makecell{84.4\\$\pm$0.49} & \makecell{83.2\\$\pm$0.4} \\ 
\cline{2-7}
& LR & N/A & \makecell{84.78\\$\pm$0.28} & \makecell{84.2\\$\pm$0.4} & \makecell{84.8\\$\pm$0.4} & \makecell{84.2\\$\pm$0.4} \\ 
\cline{3-7}
& FOLD-SE[LR] & \makecell{94.03\\$\pm$0.36} & \makecell{83.92\\$\pm$0.32} & \makecell{83.0\\$\pm$0.0} & \makecell{83.8\\$\pm$0.40} & \makecell{83.4\\$\pm$0.49} \\ 
\hline

\multirow{8}{*}{\makecell{German}} &DNN & N/A & \makecell{74.5\\$\pm$1.67} & \makecell{73.0\\$\pm$2.19} & \makecell{74.4\\$\pm$1.74} & \makecell{73.0\\$\pm$2.19} \\ 
\cline{3-7}
&FOLD-SE[DNN]& \makecell{81.5\\$\pm$2.32} & \makecell{71.6\\$\pm$1.5} & \makecell{71.0\\$\pm$2.37} & \makecell{71.6\\$\pm$1.5} & \makecell{70.8\\$\pm$1.72} \\ 
\cline{2-7}
&GBC & N/A & \makecell{75.8\\$\pm$1.63} & \makecell{74.6\\$\pm$1.62} & \makecell{76.0\\$\pm$1.67} & \makecell{74.6\\$\pm$1.62} \\
\cline{3-7}
&FOLD-SE[GBC]& \makecell{81.60\\$\pm$5.42} & \makecell{72.6\\$\pm$3.64} & \makecell{72.0\\$\pm$2.9} & \makecell{72.6\\$\pm$3.61} & \makecell{71.0\\$\pm$3.03} \\ 
\cline{2-7}
&RF & N/A & \makecell{75.7\\$\pm$1.21} & \makecell{74.6\\$\pm$2.15} & \makecell{76.0\\$\pm$1.41} & \makecell{72.8\\$\pm$1.17} \\ 
\cline{3-7}
&FOLD-SE[RF]& \makecell{85.1\\$\pm$2.37} & \makecell{71.6\\$\pm$0.20} & \makecell{70.2\\$\pm$1.94} & \makecell{71.2\\$\pm$0.40} & \makecell{66.2\\$\pm$2.86} \\ 
\cline{2-7}
&LR & N/A & \makecell{74.5\\$\pm$1.18} & \makecell{73.0\\$\pm$1.26} & \makecell{74.6\\$\pm$1.36} & \makecell{73.4\\$\pm$1.02} \\
\cline{3-7}
&FOLD-SE[LR]& \makecell{82.5\\$\pm$2.41} & \makecell{72.2\\$\pm$2.01} & \makecell{71.4\\$\pm$2.87} & \makecell{72.0\\$\pm$2.19} & \makecell{71.8\\$\pm$2.48} \\ 
\hline

\multirow{8}{*}{Car} &DNN & N/A & \makecell{94.4\\$\pm$0.31} & \makecell{97.6\\$\pm$0.49} & \makecell{97.2\\$\pm$0.40} & \makecell{97.2\\$\pm$0.4} \\ 
\cline{3-7}
&FOLD-SE[DNN] & \makecell{91.55\\$\pm$3.95} & \makecell{91.6\\$\pm$4.00} & \makecell{93.6\\$\pm$2.5} & \makecell{91.6\\$\pm$3.88} & \makecell{91.8\\$\pm$3.87} \\ 
\cline{2-7}
&GBC  & N/A & \makecell{97.5\\$\pm$1.12} & \makecell{97.4\\$\pm$1.02} & \makecell{97.4\\$\pm$1.02} & \makecell{97.4\\$\pm$1.02} \\
\cline{3-7}
&FOLD-SE[GBC] & \makecell{97.16\\$\pm$3.8} & \makecell{95.24\\$\pm$4.3} & \makecell{96.4\\$\pm$2.8} & \makecell{95.4\\$\pm$4.27} & \makecell{95.4\\$\pm$4.27} \\ 
\cline{2-7}

&RF  & N/A & \makecell{95.71\\$\pm$0.56} & \makecell{95.6\\$\pm$0.8} & \makecell{95.6\\$\pm$0.8} & \makecell{95.6\\$\pm$0.8} \\ 
\cline{3-7}
&FOLD-SE[RF] & \makecell{94.27\\$\pm$3.27} & \makecell{95.08\\$\pm$2.92} & \makecell{96.0\\$\pm$2.19} & \makecell{95.2\\$\pm$3.12} & \makecell{95.4\\$\pm$2.87} \\ 
\cline{2-7}
&LR  & N/A & \makecell{94.79\\$\pm$1.43} & \makecell{95.0\\$\pm$1.79} & \makecell{94.8\\$\pm$1.72} & \makecell{94.8\\$\pm$1.72} \\ 
\cline{3-7}
&FOLD-SE[LR] & \makecell{95.37\\$\pm$1.21} & \makecell{94.32\\$\pm$1.54} & \makecell{94.8\\$\pm$1.17} & \makecell{94.4\\$\pm$1.5} & \makecell{94.4\\$\pm$1.5} \\ 
\hline

\multirow{8}{*}{\makecell{Mushroom}} &DNN & N/A & \makecell{100.0\\$\pm$0.0} & \makecell{100.0\\$\pm$0.0} & \makecell{100.0\\$\pm$0.0} & \makecell{100.0\\$\pm$0.0} \\ 
\cline{3-7}
&FOLD-SE[DNN] & \makecell{99.57\\$\pm$0.26} & \makecell{99.56\\$\pm$0.25} & \makecell{99.4\\$\pm$0.49} & \makecell{99.4\\$\pm$0.49} & \makecell{99.4\\$\pm$0.49} \\ 
\cline{2-7}
&GBC  & N/A & \makecell{99.95\\$\pm$0.05} & \makecell{100.0\\$\pm$0.0} & \makecell{100.0\\$\pm$0.0} & \makecell{100.0\\$\pm$0.0} \\
\cline{3-7}
&FOLD-SE[GBC] & \makecell{99.57\\$\pm$0.3} & \makecell{99.56\\$\pm$0.25} & \makecell{99.4\\$\pm$0.49} & \makecell{99.4\\$\pm$0.49} & \makecell{99.4\\$\pm$0.49} \\ 
\cline{2-7}
&RF  & N/A & \makecell{100.0\\$\pm$0.0} & \makecell{100.0\\$\pm$0.0} & \makecell{100.0\\$\pm$0.0} & \makecell{100.0\\$\pm$0.0} \\ 
\cline{3-7}
&FOLD-SE[RF] & \makecell{99.57\\$\pm$0.26} & \makecell{99.56\\$\pm$0.25} & \makecell{99.4\\$\pm$0.49} & \makecell{99.4\\$\pm$0.49} & \makecell{99.4\\$\pm$0.49} \\ 
\cline{2-7}
&LR  & N/A & \makecell{99.96\\$\pm$0.05} & \makecell{100.0\\$\pm$0.0} & \makecell{100.0\\$\pm$0.0} & \makecell{100.0\\$\pm$0.0} \\
\cline{3-7}
&FOLD-SE[LR] & \makecell{99.6\\$\pm$0.27} & \makecell{99.56\\$\pm$0.25} & \makecell{99.4\\$\pm$0.49} & \makecell{99.4\\$\pm$0.49} & \makecell{99.4\\$\pm$0.49} \\ 
\hline

\end{tabular}
\caption{Performance of FOLD-SE in learning the underlying logic of various models.}
\label{tab:model_performance}
\end{table}

\section{Experiments}\label{sec_experiments}
\vspace{-0.1 in}

We have split the tasks into two steps: \textbf{1)} Find the rule based equivalent of the original model; and \textbf{2)} Using the rules obtained, find the path to the counterfactual. Further details on the experiments as well as the implementation are provided in the supplement.

\vspace{-0.07 in}
\subsection{Approximation of Models}

We demonstrate the utility of learning the underlying rules of any classifier using \textit{RBML} algorithms. For our \textit{CoGS} framework, we applied the FOLD-SE algorithm various classifiers, including \textit{Deep Neural Networks (DNN)}, \textit{Gradient Boosted Classifiers (GBC)}, \textit{Random Forest Classifiers (RF)}, and \textit{Logistic Regression Classifiers (LR)}. These classifiers are first trained on several datasets: the \textit{Adult} dataset by \cite{adult}, the \textit{Statlog (German Credit)} dataset by \cite{german}, the \textit{Car evaluation} dataset by \cite{car}, and the \textit{Mushroom} dataset by \cite{mushroom}. 
The results of these classifiers and their corresponding FOLD-SE equivalent are shown in the Table \ref{tab:model_performance}. The \textit{Accuracy (Acc.), Precision (Prec.), Recall (Rec.)} and \textit{F1 Score (F1.)} performance on the test set are shown along with the \textit{Fidelity (Fid.)} of the FOLD-SE approximation. \textit{Fidelity}, as used by \cite{ref_clear}, refers to the degree to which the \textit{RBML} algorithm (FOLD-SE) approximates the behaviour of the underlying model. A higher \textit{fidelity} score indicates that FOLD-SE closely matches the predictions of the underlying model.  

FOLD-SE successfully learns the underlying logic of these models, and the decision rules it generates (denoted as $Q$ in Definition \ref{defn:S_Q}) are used to produce counterfactuals.
While there is a penalty in performance compared to the original models, this is to be expected as FOLD-SE and other \textit{RBML} algorithms have the added constraint of being explainable and interpretable in addition to learning the underlying logic
when compared to the underlying models which may not have such constraints. For further details please see the supplement.
Next we show how the decision rules learned by our \textit{RBML algorithm} (FOLD-SE) are used to generate a path to the counterfactual.

\subsection{Obtaining a Path to the Counterfactual}

We applied the \textit{CoGS} methodology to rules generated by FOLD-SE across the \textit{German, Adult}, \textit{and Car Evaluation} datasets. These datasets include demographic and decision labels such as credit risk (`\textit{good}' or `\textit{bad}'), income (‘$=<\$50k/year$’ or ‘$>\$50k/year$’), and used car acceptability. We relabeled the car evaluation dataset -\textit{`acceptable'} or \textit{`unacceptable'}- to generate the counterfactuals. For the German dataset, \textit{CoGS} identifies paths that convert a `bad' credit rating to `\textit{good}' to determine the criteria for a favourable credit risk, i.e., likely to be approved for a loan. Similarly, \textit{CoGS} identifies paths for converting the undesired outcomes in the \textit{Adult} and \textit{Car Evaluation} datasets-‘$=<\$50k/year$’ and \textit{`unacceptable'}-to their counterfactuals: ‘$>\$50k/year$’ and \textit{`acceptable'}. We use \textit{CoGS} to obtain counterfactual paths. Details are provided in the supplement.




\begin{table}[h!]
\centering
\fontsize{10}{10}\selectfont
\begin{tabular}{|@{}c@{}|@{}c@{}|c|}
\hline
\textbf{ Dataset } & \textbf{ Avg. \# Feature Values} & \textbf{ Time (s) } \\ \hline
\multirow{2}{*}{Adult}  & 2  & $1.29 \pm 0.43$  \\ \cline{2-3} 
                        & 2.667 & $4.34 \pm 3.60$ \\ \hline
\multirow{2}{*}{German} & 2.2 & $8.98 \pm 6.257$ \\ \cline{2-3} 
                        & 2.6 & $27.82 \pm 21.72$ \\ \hline
\multirow{2}{*}{Car}    & 2   & $1.29 \pm 0.43$ \\ \cline{2-3} 
                        & 3   & $15.49 \pm 7.72$ \\ \hline
\end{tabular}
\caption{Scalability of \textit{CoGS}: Time taken to generate a counterfactual path increases as \textit{Average Feature Values} increases.}
\label{tab:scalability}
\end{table}
\textit{CoGS} can be computationally heavy as it searches through an ever-increasing solution space to find a path to a counterfactual. We run an experiment to check the effect of increasing the search space of feature values on the time taken to obtain a counterfactual solution path. 
There are two ways to increase the search space: increase the number of features or, for a fixed set of features, increase the available values that can be assigned to them. Since our decision rules are static, so are our features. Hence, we increase the number of values available to the features and test its effect on the time taken.
We only consider categorical features. This is because numeric features are handled as a range and are computationally treated as holding only one value ---the range---hence, they have a limited effect on the feature space. We use a metric \textit{Average \# Feature Values =  (Total Count of Categorical Feature Values) $\div$ (Total Count of Categorical Features)}. Table \ref{tab:scalability} shows how increasing the \textit{Average Feature Values} increases the time to find a solution. Further details of the experiment and counterfactual paths generated are available in the supplement.

\section{Conclusion and Future Work}


The main contribution of this paper is the \textit{CoGS} framework, which automatically generates paths to causally consistent counterfactuals for any machine learning model---statistical or rule-based. \textit{CoGS} efficiently computes counterfactuals, even for complex models, and the paths can be minimal if desired. By incorporating Answer Set Programming (ASP), \textit{CoGS} ensures each counterfactual is causally compliant and delivered through a sequence of actionable interventions, making it more practical for real-world applications compared to methods by \cite{ref_asp_cf} and \cite{ref_4_karimi_2}. Inspired by \cite{parth_nesy}, future work will explore improving scalability and extending \textit{CoGS} to non-tabular data, such as image classification tasks.

\bibliographystyle{apalike}
\bibliography{sample_paper}

\begin{thebibliography}{}

\bibitem[Arias et~al., 2018]{scasp-iclp2018}
Arias, J., Carro, M., Salazar, E., Marple, K., and Gupta, G. (2018).
\newblock {Constraint Answer Set Programming without Grounding}.
\newblock {\em TPLP}, 18(3-4):337--354.

\bibitem[Baral, 2003]{baral}
Baral, C. (2003).
\newblock {\em Knowledge representation, reasoning and declarative problem solving}.
\newblock Cambridge University Press.

\bibitem[Becker and Kohavi, 1996]{adult}
Becker, B. and Kohavi, R. (1996).
\newblock {Adult}.
\newblock UCI Machine Learning Repository.
\newblock {DOI}: https://doi.org/10.24432/C5XW20.

\bibitem[Bertossi and Reyes, 2021]{ref_asp_cf}
Bertossi, L.~E. and Reyes, G. (2021).
\newblock Answer-set programs for reasoning about counterfactual interventions and responsibility scores for classification.
\newblock In {\em Proc. ILP}, LNCS.

\bibitem[Bohanec, 1997]{car}
Bohanec, M. (1997).
\newblock {Car Evaluation}.
\newblock UCI Machine Learning Repository.
\newblock {DOI}: https://doi.org/10.24432/C5JP48.

\bibitem[Brewka et~al., 2011]{cacm-asp}
Brewka, G., Eiter, T., and Truszczynski, M. (2011).
\newblock Answer set programming at a glance.
\newblock {\em Commun. {ACM}}, 54(12):92--103.

\bibitem[Byrne, 2019]{ref_Byrne_CF}
Byrne, R. M.~J. (2019).
\newblock Counterfactuals in explainable artificial intelligence {(XAI):} evidence from human reasoning.
\newblock In {\em Proc. IJCAI}, pages 6276--6282.

\bibitem[Gelfond and Kahl, 2014]{gelfond-kahl}
Gelfond, M. and Kahl, Y. (2014).
\newblock {\em Knowledge Representation, Reasoning, and the Design of Intelligent Agents: The Answer-Set Programming Approach}.
\newblock Cambridge University Press, USA.

\bibitem[Hofmann, 1994]{german}
Hofmann, H. (1994).
\newblock {Statlog (German Credit Data)}.
\newblock UCI Machine Learning Repository.
\newblock {DOI}: https://doi.org/10.24432/C5NC77.

\bibitem[Karimi et~al., 2020]{alt_karimi}
Karimi, A., Barthe, G., Balle, B., and Valera, I. (2020).
\newblock Model-agnostic counterfactual explanations for consequential decisions.
\newblock In {\em AISTATS}. {PMLR}.

\bibitem[Karimi et~al., 2021]{ref_4_karimi_2}
Karimi, A., Sch{\"{o}}lkopf, B., and Valera, I. (2021).
\newblock Algorithmic recourse: from counterfactual explanations to interventions.
\newblock In {\em Proc. ACM FAccT}, pages 353--362.

\bibitem[Padalkar et~al., 2024]{parth_nesy}
Padalkar, P., Wang, H., and Gupta, G. (2024).
\newblock Ne{S}y{FOLD}: {A} framework for interpretable image classification.
\newblock In {\em Proc. AAAI}.

\bibitem[Pearl, 2009]{SCM}
Pearl, J. (2009).
\newblock {Causal inference in statistics: An overview}.
\newblock {\em Statistics Surveys}, 3(none):96 -- 146.

\bibitem[Russell, 2019]{ref_mace_2}
Russell, C. (2019).
\newblock Efficient search for diverse coherent explanations.
\newblock In {\em Proc. ACM FAT}, page 20–28.

\bibitem[Schlimmer, 1981]{mushroom}
Schlimmer, J. (1981).
\newblock {Mushroom}.
\newblock UCI Machine Learning Repository.
\newblock {DOI}: https://doi.org/10.24432/C5959T.

\bibitem[Tolomei et~al., 2017]{ref_mace_1}
Tolomei, G., Silvestri, F., Haines, A., and Lalmas, M. (2017).
\newblock Interpretable predictions of tree-based ensembles via actionable feature tweaking.
\newblock In {\em Proc. ACM SIGKDD}.

\bibitem[Ustun et~al., 2019]{ref_2_ustun}
Ustun, B., Spangher, A., and Liu, Y. (2019).
\newblock Actionable recourse in linear classification.
\newblock In {\em FAT}. {ACM}.

\bibitem[Wachter et~al., 2018]{wachter}
Wachter, S., Mittelstadt, B., and Russell, C. (2018).
\newblock Counterfactual explanations without opening the black box: Automated decisions and the gdpr.

\bibitem[Wang and Gupta, 2024]{foldse}
Wang, H. and Gupta, G. (2024).
\newblock {FOLD-SE:} an efficient rule-based machine learning algorithm with scalable explainability.
\newblock In {\em Proc. PADL 2024}, pages 37--53.
\newblock Springer LNCS 14512.

\bibitem[White and d'Avila Garcez, 2020]{ref_clear}
White, A. and d'Avila Garcez, A.~S. (2020).
\newblock Measurable counterfactual local explanations for any classifier.
\newblock In {\em Proc. ECAI}, volume 325, pages 2529--2535.

\end{thebibliography}

\end{document}